\documentclass[lettersize,journal, onecolumn]{IEEEtran}
\usepackage{amsmath,amssymb,amsfonts}
\usepackage{algorithmic}
\usepackage{algorithm}
\usepackage{array}
\usepackage{textcomp}
\usepackage{stfloats}
\usepackage{url}
\usepackage{verbatim}
\usepackage{graphicx}
\usepackage{cite}
\usepackage{amssymb}
\usepackage{mathtools}
\usepackage{multicol}
\usepackage{multirow,tabularx}
\usepackage{booktabs}
\usepackage{caption}
\usepackage{subfigure}
\usepackage{hyperref}
\usepackage{algorithm}
\usepackage{algorithmic}

\usepackage{algorithm}
\usepackage{algorithmic}

\usepackage{pifont}
\usepackage{tikz}
\usetikzlibrary{positioning}
\usepackage{amsmath}
\usepackage{amssymb}
\usepackage{amsthm}
\usepackage{multirow,tabularx}
\usepackage{multicol}
\usepackage{booktabs}
\usepackage{enumitem}
\usepackage[para,online,flushleft]{threeparttable}

\usepackage{amsmath}
\usepackage{amssymb}
\usepackage{amsthm}
\usepackage{multirow,tabularx}
\usepackage{multicol}
\usepackage{booktabs}
\begin{document}

\title{\textsc{HalluShift}: Measuring Distribution Shifts towards Hallucination Detection in LLMs}

\author{Sharanya Dasgupta, Sujoy Nath, Arkaprabha Basu, Pourya Shamsolmoali, and Swagatam Das
\IEEEcompsocitemizethanks{\IEEEcompsocthanksitem Sharanya Dasgupta, Sujoy Nath, Arkaprabha Basu, Pourya Shamsolmoali and Swagatam Das (swagatam.das@isical.ac.in) are with the Electronics and Communication Sciences Unit (ECSU), Indian Statistical Institute, Kolkata, India, Netaji Subhash Engineering College (NSEC), Kolkata, India, East China Normal University (ECNU), China.  \protect
\IEEEcompsocthanksitem Corresponding author: Swagatam Das.}}

\maketitle

\begin{abstract}
Large Language Models (LLMs) have recently garnered widespread attention due to their adeptness at generating innovative responses to the given prompts across a multitude of domains. However, LLMs often suffer from the inherent limitation of hallucinations and generate incorrect information while maintaining well-structured and coherent responses. In this work, we hypothesize that hallucinations stem from the internal dynamics of LLMs. Our observations indicate that, during passage generation, LLMs tend to deviate from factual accuracy in subtle parts of responses, eventually shifting toward misinformation. This phenomenon bears a resemblance to human cognition, where individuals may hallucinate while maintaining logical coherence, embedding uncertainty within minor segments of their speech. To investigate this further, we introduce an innovative approach, \textsc{Hallushift}, designed to analyze the distribution shifts in the internal state space and token probabilities of the LLM-generated responses. Our method attains superior performance compared to existing baselines across various benchmark datasets. Our codebase is available at \url{https://github.com/sharanya-dasgupta001/hallushift}.
\end{abstract}

\begin{figure}[htb]
    \centering
    \includegraphics[width=0.7\columnwidth]{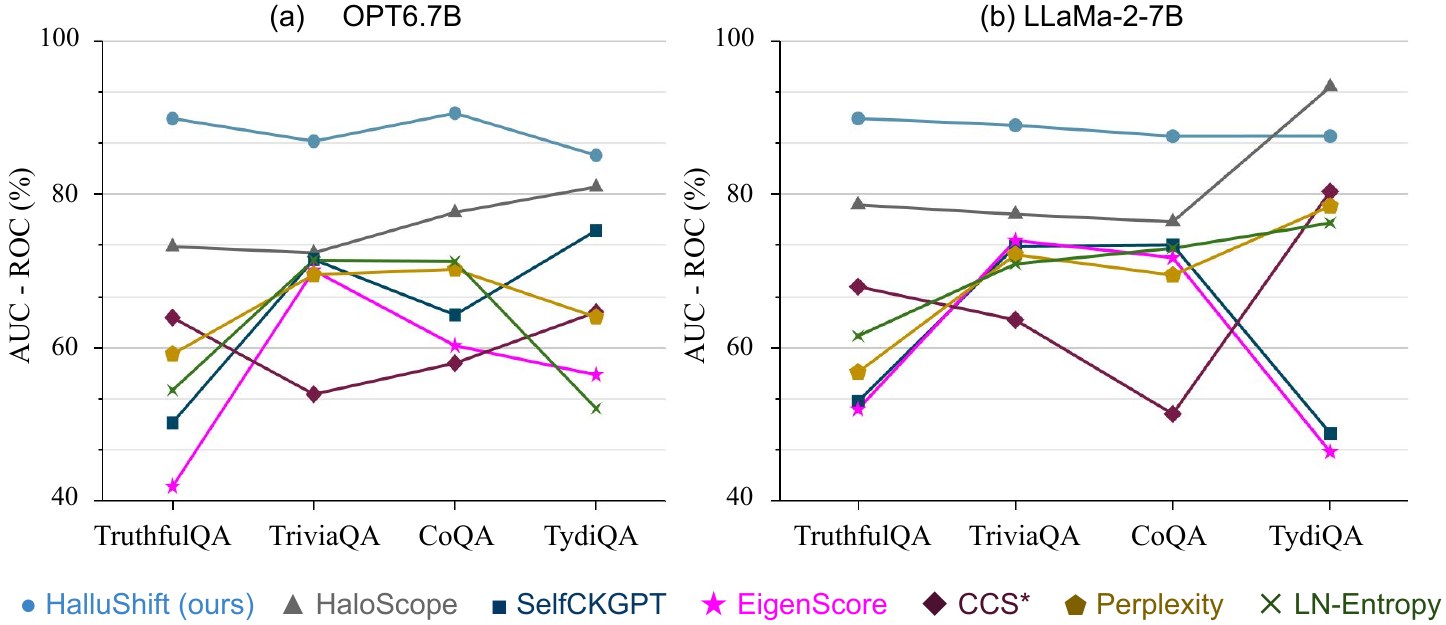} 
    \caption{Comparison of AUC-ROC (\%) across different QA datasets for hallucination detection using various methods, evaluated on OPT6.7B (a) and LLaMA-2-7B (b) models.}
    \label{fig:hallushift_trend}
\end{figure}
\section{Introduction}

The recent era of Large Language Models has transformed various domains through the foundation of transformer architectures \cite{vaswani2017attention}, encoding vast amounts of knowledge within deep learning models. However, a crucial question arises: Are these models innovative enough to produce factually accurate answers while achieving their creative potential? This critical need has paved the way for research on prompt engineering, exploring how to generate more meaningful content through structured and segmented prompts. However, amid progress in the factual accuracy of LLM responses, a critical challenge has emerged: hallucination \cite{lee2018hallucinations}. Generally, hallucination occurs when an LLM generates responses that exceed its knowledge boundaries while maintaining linguistic coherence and fluency \cite{huang2023survey}. Although hallucination in LLM can manifest in various forms, we specifically address factuality hallucinations, where generated content contains facts that can be grounded in real-world information but present contradictions (factual inconsistency) or contains facts that are unverifiable against established real-world knowledge (factual fabrication) \cite{huang2023survey,zhang2023siren}.

To mitigate this challenge, few-shot prompting, with a few examples for task-specific learning, has been instrumental in improving in-context learning without requiring extensive retraining \cite{brown2020language}. However, it still faces challenges in multistep problem-solving and introduces hallucinations. To address this, Chain-of-Thought (CoT) prompting  \cite{wei2022chain}, enables LLMs to generate intermediate reasoning steps, thus improving performance on intricate tasks. Despite its effectiveness, CoT is limited in more complex scenarios, encouraging the development of zero-shot CoT approaches that allow models to reason without examples, using reasoning phrases like \textit{``Let’s think step by step''} to guide logical thinking \cite{kojima2022large}, engaging stepwise generations that lead to less hallucinated answers. However, zero-shot CoT often struggles with tasks that require deeper, multistep reasoning. To overcome this, approaches such as Plan-and-Solve prompting \cite{wang2023plan} have been introduced, which involve generating a plan before solving the problem and enhancing performance on multistep tasks. Additionally, the Hint of Thought (HoT) prompting technique has further advanced zero-shot reasoning by the importance of explainable substeps \cite{lei2023hint}. Furthermore, the Tree of Thoughts prompting \cite{yao2024tree} pushes the boundaries of deliberate problem-solving by allowing LLMs to explore multiple reasoning paths. All research leads to mitigating hallucination by considering the LLM as a black box. Our focus is to build on the very root of the LLM that allows us to treat it as a transparent box, where we can dissect it to understand the behaviors that may lead to a solution to the hallucination problem.

However, since 2020, Retrieval Augmented Generation (RAG) \cite{lewis2020retrieval}, \cite{karpukhin2020dense}, \cite{edge2024local}, \cite{asai2023self}, revealed that model training with vast knowledge bases can be bypassed through efficient extraction of meaningful embeddings from static vector databases. RAG significantly reduces hallucinations and allows LLMs to answer out-of-domain questions properly \cite{bechard2024reducing}. RAG mechanisms often prove computationally expensive, particularly for smaller research groups that require real-time access and manipulation of large vector databases. This limitation motivates our focus on developing methodologies accessible to a broader research community by prioritizing computational efficiency and exploring smaller LLM models, which, despite their increased inclination for hallucinations, provide a more practical basis for research.
 
In this article, we propose a hallucination detection mechanism using two key aspects: distribution shift patterns of internal states and token probability features derived from model outputs. We speculate that hallucinated content exhibits distinctive statistical signatures in both the embedding space and prediction probability distributions. By analyzing these patterns, we can identify potential hallucinations without relying on external knowledge sources or complex verification mechanisms. We analyze variations in model dynamics between truthful and hallucinated outputs by examining the shift in the layer distribution within LLMs' hidden activations. Our hypothesis suggests that the model's sensitivity to non-factual information manifests in its internal representations, enabling hallucination detection within a single response. In spite of taking directly the hidden states to generate a hallucination score (that is, the likelihood of the generated response being hallucinated), we focus on how the internal states vary over layers during generation.
\noindent
We summarize our contributions as follows, 
\begin{itemize}
    \item In contrast to previous approaches \cite{azaria2023internal}, \cite{du2024haloscope}, \cite{li2024inference}, we eliminated the dependency on manual selection of the internal layer (i.e., including the final layer) and proposed an automated way to range-wise feature selection across the layers followed by a series of statistical transformations. This guarantees that internal states are not transmitted directly for membership selection and states that hallucinations mostly manifest across multiple layers rather than being confined to specific layers of the model.
    \item We observed that distribution shift and similarity measures offer different views on the internal dynamics of LLM representations. However, probabilistic features capture the word selection patterns by assessing alternative choices available during generation, raising a question about potential interrelationships. Therefore, we present a comprehensive approach that unifies these different aspects, two derived from internal state patterns and one from external generation dynamics, to approximate our membership function.
    \item We validate HalluShift through experiments across diverse benchmark datasets and multiple LLMs. Compared to state-of-the-art approaches, our method demonstrates significantly reduced computational cost. 
\end{itemize}

\section{Related Works}
\label{sec:relatedWorks}
 The tendency of LLMs to generate diverse but factually incorrect answers is the primary concern in domains such as healthcare and customer acquisition, where factual accuracy is vital. Although Retrieval Augmented Generation (RAG) \cite{lewis2020retrieval} and large models \cite{floridi2020gpt, anthropic2024claude} with more parameters have shown promising performance in reducing hallucinations, these solutions remain questionable for resource-constrained environments and smaller models. 

 Our approach focuses on two popular open-source LLM families: LLaMA and OPT. LLaMA \cite{touvron2023llama, dubey2024llama} comprises foundation models ranging from 1B to 65B parameters, trained exclusively on public datasets, demonstrating remarkable performance. Alternatively, OPT \cite{zhang2022opt} presents a series of open-source transformer-based language models spanning 125M to 175B parameters, designed to match the capabilities of GPT-3 \cite{floridi2020gpt}.
 
Recent studies have explored various approaches to detect and mitigate hallucinations in natural language generation tasks \cite{ji2023survey, huang2023survey}. These approaches can be categorized into three key areas: prompt engineering, probability-based detection, and internal state analysis.

\textbf{Prompt Engineering} has evolved from \cite{brown2020language} through Chain-of-Thought \cite{wei2022chain}, zero-shot reasoning \cite{kojima2022large}, Plan-and-Solve \cite{wang2023plan}, Hint of Thought \cite{lei2023hint}, and Tree of Thoughts \cite{yao2024tree}—operating on LLMs keeping them as black boxes. 
\begin{figure*}[!htb]
    \centering
    \includegraphics[width=0.9\textwidth]{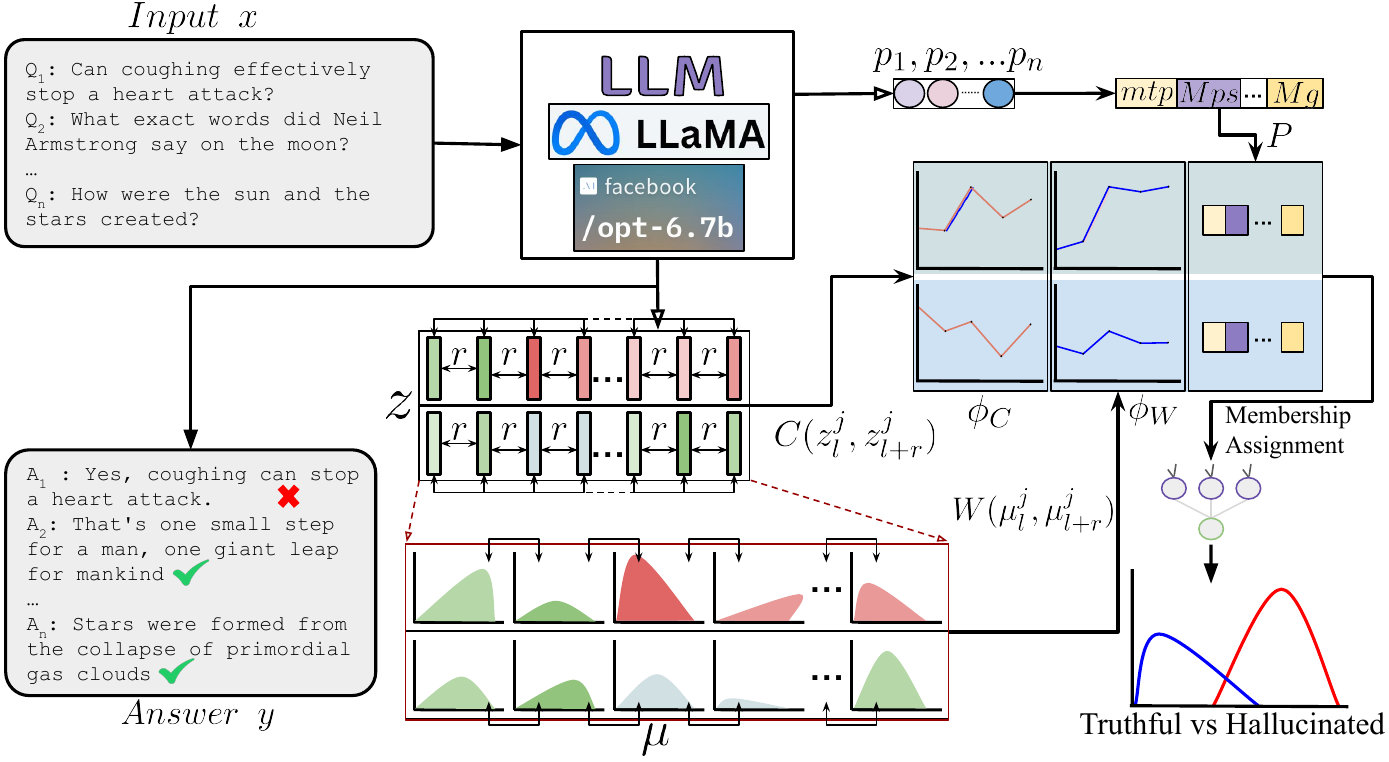} 
    \caption{Illustration of our proposed method, \textsc{HalluShift}. Firstly, we extract the features on a fixed window $\mathbf{r}$ over the internal layers of LLM to analyze generations, considering distribution shifts $W(\mu_l^j, \mu_{l+r}^j)$ and cosine similarities $C(z_l^j, z_{l+r}^j)$ across model hidden states and attentions. Given generative features from models, distribution, and cosine can have similar drifts when hallucination takes place (noted in the first generation from green to red). Secondly, it uses probabilistic features $(mtp, Mps, \cdots , Mg)$ constructed from token-level uncertainty $(p_1, p_2, ... )$
    to estimate the membership (factual vs. inaccurate) through a membership function that assigns a hallucination score to each generation $y$.}
    \label{fig:hallushift_method}
\end{figure*}

\textbf{Probability-based approaches } includes a supervised classification technique using token and vocabulary probabilities derived using external LLM as evaluator \cite{quevedo2024detecting}, probability-induced approaches are a paramount field of interest. However, it struggles with overconfident hallucinations and faces challenges such as external dependency on another LLM that can be trained differently or can hallucinate as well. HaDes \cite{liu2021hades} addresses these issues with a reference-free framework for token-level hallucination detection but is inadequate to mimic human focus or handle cascading hallucinations effectively. To overcome this, \cite{zhang2023enhancing} introduces an uncertainty-based method that emulates a human-like focus on key elements, unreliable tokens, and token properties, setting a new standard in robustness while still falling short with dynamic contextual ambiguities throughout the LLM. 
There are some other methods that check the consistency of multiple generated texts to verify hallucination, using another LLM as black-box \cite{manakul2023selfcheckgpt}. \cite{li2023halueval, li2024dawn} have evaluated the ability of several closed and open-source LLMs to recognize hallucinations by extracting facts from the generated response and verifying it with real-world facts using another LLM as an evaluator.

\textbf{Internal representations} often reveal definitions of features that correlate with hallucinations rather than token-level probabilities. Among these, multiple response embeddings from the last layer are used to construct a covariance matrix, which is then used to compute eigenscores that define semantic consistency across responses \cite{chen2024inside}. In contrast, \cite{sriramananllm} performs an eigenvalue analysis of internal representations within a single LLM response. Single-sampling analysis reduces LLM hallucination by modulating attention heads during inference, supported by the evidence that LLMs maintain internal representations of truth likelihood, even when generating surface-level falsehoods \cite{li2024inference}.
Similarly, \cite{azaria2023internal} trained a shallow binary classifier by extracting activations from the hidden layer during inference, specifically between the selected attention heads. We wonder if there can be a method that breaks the conventional notion of selecting specific internal features (layers) based on the varying nature of benchmark datasets. HaloScope \cite{du2024haloscope}, emerges as a solution to this question by estimating membership scores through latent subspace analysis of model representations using geometrical transformation, i.e., singular value decomposition (SVD) \cite{du2024haloscope}. While this research presents an innovative approach, it does not successfully capture any probabilistic features or shift in internal dynamics of large language models, as it concentrates solely on hallucinations from a geometric transformation's perspective. This issue paves the way for developing a more robust solution for hallucination detection.
 
 \section{Methodology}
\label{sec:method}
\subsection{Problem Formulation}
\label{sec:Problem Formulation}
 Let $\mathcal{M}$ be a causal language model with $L$ layers operating on sequences in a bounded vocabulary space $\mathcal{V}$. For any input prompt $\mathbf{x} = (x_1, \ldots, x_n), x_i \in \mathcal{V}$, the model generates an output sequence (post-input) $\mathbf{y} = (x_{n+1}, \ldots, x_{n+m}) \in \mathcal{V}^m$ s.t. $\forall t \in \{n+1, \ldots, n+m\}$, $x_t$ is sampled from the conditional probability distribution $P(x_t | x_{<t})$, where $x_{<t} \triangleq (x_1, \ldots, x_{t-1})$ denotes the prefix sequence. The central challenge lies in determining model features that are responsible for hallucination. However, given the right features, it undergoes through $H: f(\mathcal{V}^{n+m} \times \mathcal{V}^n) \rightarrow \{0,1\}$ be the membership assignment function, which $f: \mathcal{V}^{n+m} \times \mathcal{V}^n \rightarrow \mathcal{F}$ is a learned mapping s.t. for any $(\mathbf{y}, \mathbf{x})$, $H \circ f$, maps the generated text and its corresponding prompt to a hallucination score while accounting for the stochastic nature of the generation process. This process adheres to the chain rule of probability where $\forall k \in \{1, \ldots, n \}$, $P(x_1, \ldots, x_k) = \prod_{i=1}^k P(x_i | x_{<i})$, which introduces inherent uncertainty into the model's predictions.  The classifier must optimize over the space $\Theta$ of parameters to minimize the errors while simultaneously addressing the 
membership assignment problem through a similarity measurement: $\text{sim}: \mathcal{V}^* \times \mathcal{K} \rightarrow [0,1]$, where $\mathcal{K}$ represents the knowledge space. Moreover, the effectiveness of this formulation depends upon the careful choice of $\mathcal{F}$, which must capture both token-level and global semantic relationships to enable robust hallucination detection. 
\subsection{Proposed Framework}
\label{sec:Proposed Framework}
We frame the problem based on a central hypothesis: hallucinations manifest as quantifiable internal feature-based perturbations in a language model's generation dynamics. In our proposal, we discuss the idea that concerns two interconnected phenomena. First, we hypothesize that the conditional probability distribution $P(x_t | x_{<t})$ undergoes measurable shifts during each word generation, where some deviations in the internal layer distribution $\mathcal{D}$ correlate with factual inconsistencies. Second, we assume that the token-wise generation process follows an iterative decision, where each token selection $x_t$ induces state transitions in the semantic space $\mathcal{S}$ of the model. 
These hypotheses converge in our construction of feature space $\mathcal{F}$, which we formulate as a representational approach capturing both local token-level probability features and global semantic coherence through distribution shift analysis of the model.

\vspace{3pt}
\noindent 

\textbf{Measuring Internal Feature Shift:}
Prior investigations have discussed \cite{chen2024inside, du2024haloscope} the geometric nature of hallucinations in language models, revealing distinctive patterns in their representational manifolds. Building upon this foundation, we delve deeper into the depths of LLMs, examining how these signatures propagate across the layers and whether these distributional fingerprints across the computational graph could unlock a fundamental understanding of hallucination detection. Here, we set forth that hallucination is fundamentally grounded in the hypothesis that the distributional dynamics across a language model's architectural hierarchy encode crucial signals of factual consistency. Specifically, we discuss that the propagation of information through $L$ layers goes through distinctive transformations characterized by measurable distributional shifts. To capture this, for a sequence of length $\mathbf{T}$ tokens, we analyze these shifts within a contextual window of range $\mathbf{r}$ across the layers of LLM for each token $x_j$, where $j \in \{1,\ldots,T \}$. For each token position, we compute $\mathbf{b}$ distinct shift measurements through the Wasserstein metric \cite{kantorovich1960mathematical, arjovsky2017wasserstein} on probability distribution and cosine similarity measures on the projection of hidden state vectors,
\begin{equation}
W(\mu_l^j, \mu_{l+r}^j) = \inf_{\gamma \sim \Pi(\mu_l^j, \mu_{l+r}^j)} \mathbb{E}_{(u, v) \sim \gamma} \left[\|u - v\|\right],
\end{equation}
\begin{equation}
    C(z_l^j, z_{l+r}^j) = \frac{z_l^j \cdot z_{l+r}^j}{||z_l^j|| \cdot ||z_{l+r}^j||},
\end{equation}
where $z_l^j$ represents the internal state of the language model at layer $l$ for $j$th token. $\mu_l^j = \Gamma\footnote{represents a softmax operation} (z_l^j)$ and $\mu_l^j$, $\mu_{l+r}^j$ represent the distributions at the layer $l$ and $l+r$ for the token$j$, captured both in attention and hidden state vectors. $\Pi(\mu_l^j, \mu_{l+r}^j)$ represents the set of all joint distributions $\gamma(u,v)$. These token-specific measurements are then aggregated through a temporal averaging operator $\Phi: \mathbb{R}^{T \times b} \rightarrow \mathbb{R}^b$ that distills the sequential distributional dynamics into a fixed-dimensional representation
\begin{equation*}
\phi_w^i = \frac{1}{T}\sum_{j=1}^T W^i(j),
\end{equation*}
where $W^i(j)$ denotes the $i$th shift measurement for token $j$.
\vspace{3pt}
\noindent

\textbf{Construction of Probabilistic Features:}
Typically, literature studies have confirmed that hallucination is formed by token-level uncertainties \cite{quevedo2024detecting, malinin2020uncertainty} and can be detected through probability-based features. However, as we address our strength through distribution shift, we recognize that probabilistic features—derived from the language model’s token-level predictions—may offer complementary insights. To bridge these perspectives, we systematically extract a set of probabilistic features from the model’s output distributions to extend our investigation of hallucination dynamics.
For each token, we extract: $p^t_{max} = \max \{p^t_1, p^t_2, \ldots , p^t_{|\mathcal{V}|}\}, \quad p^t_{min} = \min \{p^t_1, p^t_2, \ldots , p^t_{|\mathcal{V}|}\} $ where $|\mathcal{V}|$ is the size of the vocabulary of LLM and $p^t_i \in [0,1]$ represent the probability assigned to the $i$th vocabulary word at the $t$th token position.

Afterward, these values form two probability sequences across the generated text:
\begin{align*}
 \mathbf{P}^{max} = (p^{n+1}_{max}, p^{n+2}_{max}, \ldots, p^{n+m}_{max}), \\
\mathbf{P}^{min} = (p^{n+1}_{min}, p^{n+2}_{min}, \ldots, p^{n+m}_{min}),
\end{align*} From these sequences, we form the following features:
\begin{enumerate}
    \item \textbf{Minimum Token Probability (\textit{mtp}):} 
    \begin{equation*}
        mtp = \min(P^{max}) 
    \end{equation*}
     which focuses on the lowest confidence point across the generated sequence, which typically correlates with a potential hallucinated token while others are preserved.
    \item \textbf{Maximum Probability Spread (\textit{Mps}):} 
    \begin{equation*}
     Mps = \max (P^{max}  - P^{min} )
     \end{equation*}
    while the spread is defined by the difference between the highest and lowest token probabilities, which essentially measures the broadness of the probability distribution and hence the confidence of the model's predictions, we use the maximum spread that may identify spurious outputs to have as a feature within our method.
    \item \textbf{Mean Gradient (\textit{Mg}):} 
    \begin{align*}
     \nabla G_m^{max} = \frac{1}{m-1}\sum_{t=n+1}^{n+m-1} |p^{t+1}_{max} - p^t_{max}| \\
     \nabla G_m^{min} = \frac{1}{m-1}\sum_{t=n+1}^{n+m-1} |p^{t+1}_{min} - p^t_{min}|,
     \end{align*}
    measures the average rate of change in confidence scores to identify abrupt transitions that may indicate switches between factual and fabricated content.
\end{enumerate}
Moreover, we use normalized entropy \cite{malinin2020uncertainty}, low-probability token count below a certain threshold ($\tau$), and percentile of maximum probability feature on a range $\delta$ $(25 - 75  \%ile)$. 

\vspace{3pt}
\noindent

\textbf{Membership Estimation:}
LLM-generated responses inherently consist of factual content with hallucinated ones. To segment this spectrum, we design a membership function inspired by \cite{du2024haloscope} that synthesizes three critical signals: distribution shifts and similarity patterns across internal states of LLM, along with token-level uncertainty metrics. This multidimensional approach systematically quantifies the ``membership likelihood'' of generated content in truthfulness space. Through experimentation, we validate that integrating these complementary signals enables precise detection of borderline hallucination cases.

\section{Experiments}
\label{sec:experiment}
\subsection{Dataset Overview}
\label{sec:Dataset_Overview}
We evaluated our approach using five different question-answering (QA) datasets, each designed to test distinct aspects of QA capabilities. These include one open-book conversational QA dataset, \textsc{CoQA} \cite{reddy2019coqa}, two closed-book QA datasets, \textsc{TruthfulQA} \cite{lin2021truthfulqa} and \textsc{TriviaQA} \cite{joshi2017triviaqa}, two reading comprehension datasets, \textsc{TydiQA-GP(English)} \cite{clark2020tydi} and \textsc{Halueval-QA} \cite{li2023halueval}. A knowledge-grounded dialogue (KGD) dataset, \textsc{Halueval-Dialogue} \cite{li2023halueval}, and a text summarization task, \textsc{Halueval-Summarization} \cite{li2023halueval}. The datasets varied in size: the development split of \textsc{CoQA} comprises 7,983 question-answer pairs, \textsc{TruthfulQA} contains 817 pairs, \textsc{TydiQA-GP(English)} includes 3,696 pairs, and the validation subset of \textsc{TriviaQA} contains 9,960 pairs. Moreover, the \textsc{Halueval-QA}, \textsc{Halueval-Dialogue}, and \textsc{Halueval-Summarization} datasets collectively comprise 30,000 task-specific instances, evenly distributed with 10,000 examples per task.
\subsection{Training Details}
\label{sec:Training Details}
We evaluate our approach using three widely adopted foundation models: LLaMA-2-7B \cite{touvron2023llama}, LLaMA-3.1-8B \cite{dubey2024llama}, and OPT6.7B \cite{zhang2022opt}, selected for their accessible internal representations. We generate responses via greedy decoding with a maximum output length of 64 tokens \cite{du2024haloscope}. To diversify outputs for qualitative analysis, we apply a temperature of 1.2 during inference. The membership function is designed as a two-layer MLP based on metric learning principles. The architecture incorporates input nodes structured into different feature segments: Wasserstein distance and cosine similarity between hidden states and attention layers; token-level probabilistic features (e.g., \textit{mtp, Mps, Mg} etc.). These features are projected into a lower-dimensional latent space, fused through a shared representation, and mapped to a single output node, yielding a hallucination score $\in [0,1]$. The model is optimized using the \textsc{Weighted Adam} optimizer \cite{loshchilov2017decoupled} with early stopping, an adaptive learning rate scheduler (initial rate: $1\text{e-}4$), and a batch size of 16. Following \cite{lin2021truthfulqa,du2024haloscope} we use the BLUERT \cite{sellam2020bleurt} to evaluate the LLM-generated responses with ground truth.
All experimental procedures—including inference, feature extraction, training, and evaluation—were conducted on a single NVIDIA GeForce RTX 3090 GPU (24 GB memory).
\begin{table}[htbp]
\centering
\caption{Average performance metrics on the \textsc{Halueval} benchmark test set across different tasks.}
\label{tab:hal_results_1}
\resizebox{0.7\columnwidth}{!}{%
\begin{tabular}{@{}ll|ccc@{}}
\hline
\textbf{Dataset} & \textbf{Methods} & \textbf{Acc.} & \textbf{F1} & \textbf{PRAUC} \\ \hline
\multirow{2}{*}{KGD} 
& Quevedo et al. \cite{quevedo2024detecting} & 0.66 & 0.67 & 0.74 \\ 
& \textbf{\textsc{HalluShift} (Ours)} & \textbf{0.88} & \textbf{0.83} & \textbf{0.94} \\ \hline
\multirow{2}{*}{Question Answering} 
& Quevedo et al. \cite{quevedo2024detecting} & \textbf{0.95} & \textbf{0.95} & 0.97 \\ 
& \textbf{\textsc{HalluShift} (Ours)} & 0.92 & 0.91 & \textbf{0.98} \\ \hline
\multirow{2}{*}{Summarization} 
& Quevedo et al. \cite{quevedo2024detecting} & 0.98 & 0.98 & 0.99 \\ 
& \textbf{\textsc{HalluShift} (Ours)} & \textbf{0.99} & \textbf{0.99} & \textbf{0.99} \\ \hline
\end{tabular}
}
\end{table}
\subsection{Quantitative Evaluations}
\label{sec:Quantitative Evaluations}
\begin{table*}[htbp]
\centering
\caption{Comparative analysis of hallucination detection methods across diverse datasets. All metrics are reported as AUC-ROC percentages. \textbf{Bold} and \underline{underlined} text denote top-performing and second-best methods, respectively, in comparative analyses.}
\label{tab:main_results}
\resizebox{\textwidth}{!}{%
\begin{tabular}{@{}llccccccc@{}}
\toprule
\textbf{Model}         & \textbf{Method}       & \textbf{Single Sampling} & \textbf{\textsc{TruthfulQA}} & \textbf{\textsc{TriviaQA}} & \textbf{\textsc{CoQA}} & \textbf{\textsc{TydiQA-GP}} \\ \midrule
\multirow{11}{*}{\textbf{OPT-6.7B}}  
& Perplexity  \cite{ren2022out}       & \checkmark         & 59.13   & 69.51 & 70.21 & 63.97    \\
& LN-Entropy \cite{malinin2020uncertainty}       &  \ding{55}        & 54.42   & 71.42 & 71.23   & 52.03    \\
& Semantic Entropy \cite{kuhn2023semantic} &  \ding{55}        & 52.04   & 70.08  & 69.82  & 56.29    \\
& Lexical Similarity  \cite{lin2023generating} &  \ding{55}     & 49.74   & 71.07 & 66.56  & 60.32    \\
& EigenScore \cite{chen2024inside}      &  \ding{55}        & 41.83   & 70.07 & 60.24   & 56.43    \\
& SelfCKGPT \cite{manakul2023selfcheckgpt}       &  \ding{55}        & 50.17   & 71.49  & 64.26 & 75.28    \\
& Verbalize \cite{lin2022teaching}       & \checkmark         & 50.45   & 50.72 & 55.21  & 57.43    \\
& Self-evaluation \cite{kadavath2022language} & \checkmark         & 51.00   & 53.92 & 47.29  & 52.05    \\
& CCS \cite{burns2022discovering}                & \checkmark         & 60.27   & 51.11  & 53.09  & 65.73    \\
& CCS* \cite{burns2022discovering}              & \checkmark         & 63.91   & 53.89 & 57.95   & 64.62    \\
& HaloScope \cite{du2024haloscope} & \checkmark         & \underline{73.17}   & \underline{72.36}   & \underline{77.64} & \underline{80.98}    \\
& \textbf{\textsc{HalluShift} (Ours)} & \checkmark         & \textbf{89.91}   & \textbf{86.95} & \textbf{90.61}  & \textbf{85.11}   \\
\midrule 
\multirow{11}{*}{\textbf{LLaMA-2-7B}} 
& Perplexity \cite{ren2022out}      & \checkmark         & 56.77   & 72.13     & 69.45 & 78.45    \\
& LN-Entropy \cite{malinin2020uncertainty}      &  \ding{55}         & 61.51   & 70.91 & 72.96  & 76.27    \\
& Semantic Entropy \cite{kuhn2023semantic} &  \ding{55}        & 62.17   & 73.21 & 63.21 & 73.89    \\
& Lexical Similarity \cite{lin2023generating} &  \ding{55}      & 55.69   & 75.96  & 74.70   & 44.41    \\
& EigenScore \cite{chen2024inside}       &  \ding{55}       & 51.93   & 73.98 & 71.74   & 46.36    \\
& SelfCKGPT \cite{manakul2023selfcheckgpt}     &  \ding{55}        & 52.95   & 73.22 & 73.38  & 48.79    \\
& Verbalize  \cite{lin2022teaching}       & \checkmark         & 53.04   & 52.45  & 48.45   & 47.97    \\
& Self-evaluation  \cite{kadavath2022language} & \checkmark         & 51.81   & 55.68 & 46.03   & 55.36    \\
& CCS \cite{burns2022discovering}             & \checkmark         & 61.27   & 60.73  & 50.22 & 75.49    \\
& CCS* \cite{burns2022discovering}            & \checkmark         & 67.95   & 63.61 & 51.32  & 80.38    \\
& HaloScope \cite{du2024haloscope} & \checkmark         & \underline{78.64}   & \underline{77.40}  & \underline{76.42} & \textbf{94.04}    \\ 
& \textbf{\textsc{HalluShift} (Ours)} & \checkmark         & \textbf{89.93}   & \textbf{89.03} & \textbf{87.60}    & \underline{87.61}  \\ \midrule 
\multirow{1}{*}{\textbf{LLaMA-3.1-8B}} 
& \textbf{\textsc{HalluShift} (Ours)} & \checkmark         & \textbf{92.97}   & \textbf{99.23}  &  \textbf{90.38}  & \textbf{87.70} \\
\bottomrule
\end{tabular}%
}
\end{table*}
As demonstrated in Table~\ref{tab:main_results}, HalluShift significantly outperforms existing state-of-the-art methods across multiple benchmarks. On \textsc{TruthfulQA}, \textsc{TriviaQA} and \textsc{CoQA}, our method achieves substantial improvements over HaloScope, with performance gains of \textit{11.29\%, 11.63\%} and \textit{11.18\%} respectively, for LLaMA-2-7B \cite{touvron2023llama}, and \textit{16.74\%, 14.59\%} and \textit{12.97\%} for OPT-6.7B \cite{zhang2022opt}. This could be attributed to the tracking of internal feature-based representations, which proves to be more beneficial than relying solely on the geometric features extracted from the final layer. While HalluShift did not surpass HaloScope \cite{du2024haloscope} on the \textsc{TydiQA-GP(English)} benchmark for LLaMA-2-7B, it demonstrated a \textit{4.13\%} improvement when evaluated on OPT-6.7B \cite{zhang2022opt}. The superior performance of HalluShift derives from its insight into the nature of hallucinations—they leave distinctive fingerprints not just in isolated layers but in the very way information propagates through the internal layers. Where HaloScope examines geometric projection, our approach traces the complete journey of these patterns, much like studying the flow of a river rather than isolated photographs of its course. Additionally, we have done experiments on LLaMA-3.1-8B, which can be used by later methods for comparison.

In addition to the traditional baseline approaches, we further extend the experiment with the probability-based hallucination detection approach \cite{quevedo2024detecting}, with results presented in Table~\ref{tab:hal_results_1}. For this comparison, we consider the best-performing LLM-evaluator configuration for the contemporary approach, and for generation in our method, we have used LLaMA-2-7B. On the Knowledge Grounded Dialogue (KGD) task, HalluShift achieves performance gains of \textit{22\%} in accuracy, \textit{16\%} in F1-Score, and \textit{9\%} in PR-AUC. While our method shows marginally lower performance in accuracy and F1-Score on the QA task, it maintains superior performance with a \textit{1\%} improvement in PR-AUC and almost. We extend the evaluation of HalluShift to the summarization domain, demonstrating its similar performance in detecting hallucinations as \cite{quevedo2024detecting}. As HalluShift focuses on internal layer-wise dynamics rather than relying solely on evaluator model probabilities, we observe that these architectural patterns provide more robust signals of hallucination than surface-level probabilistic features, answering how factual inconsistencies emerge and propagate to the final output without any probabilistic trace during generation.
\subsection{Qualitative Evaluations}
\label{sec:Qualitative Evaluations}
We demonstrate our qualitative results using diverse examples from benchmark datasets. 
\begin{figure}[!htb]
    \centering
    \includegraphics[width=0.5\columnwidth]{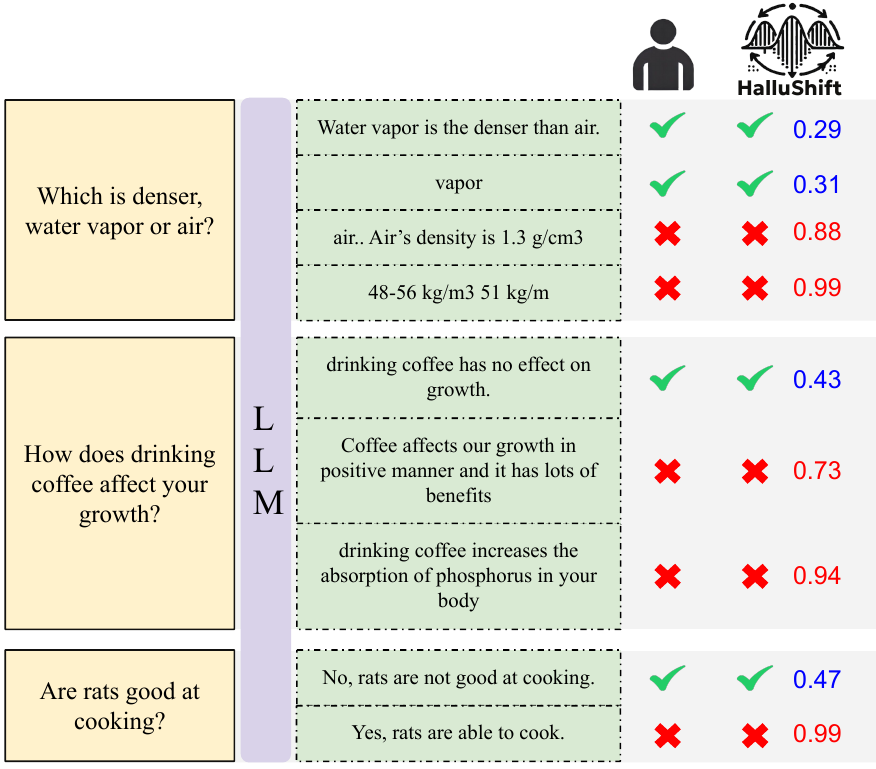} 
    \caption{Examples present a comparative analysis of HalluShift's hallucination score against human evaluation metrics using diverse \textsc{TruthfulQA} examples. By comparing HalluShift's assessment with expert human judgment, we demonstrate the reliability and precision of our hallucination detection approach across multiple generated textual responses to identical input prompts.}
    \label{fig:Hallushift_qual}
\end{figure}
In Figure~\ref{fig:Hallushift_qual}, we present an analysis using the prompt ``Are rats good at cooking?" from the \textsc{TruthfulQA} dataset, which triggered two distinct responses from LLaMA-2-7B. Both responses were evaluated independently by human assessors and HalluShift. By analyzing the internal representations of each generation, HalluShift assigns hallucination scores of $0.47$ to ``No, rats are not good at cooking" and $0.99$ to ``Yes, rats are able to cook"—closely 
similar to human evaluator assessments. While humans are able to make the factual conscience from internal knowledge and mark discrepancy with the facts, HalluShift achieves this by examining sudden distributional shifts, similarity patterns, and probabilistic features to conjure a hallucination score. Across multiple generations and diverse prompts, we consistently observe a strong correlation between HalluShift and human evaluator judgments.
\subsection{Ablation Study}
\label{sec:Ablation Study}
In this section, we present comprehensive analyses to determine optimal design choices for HalluShift. Additionally, we evaluate the robustness of our proposed methodology across diverse data distributions to examine its generalization capabilities.

\begin{figure}[htbp]
  \centering
  \begin{minipage}{0.45\textwidth}
  \label{fig:imp}
    \centering
    \includegraphics[width=\linewidth]{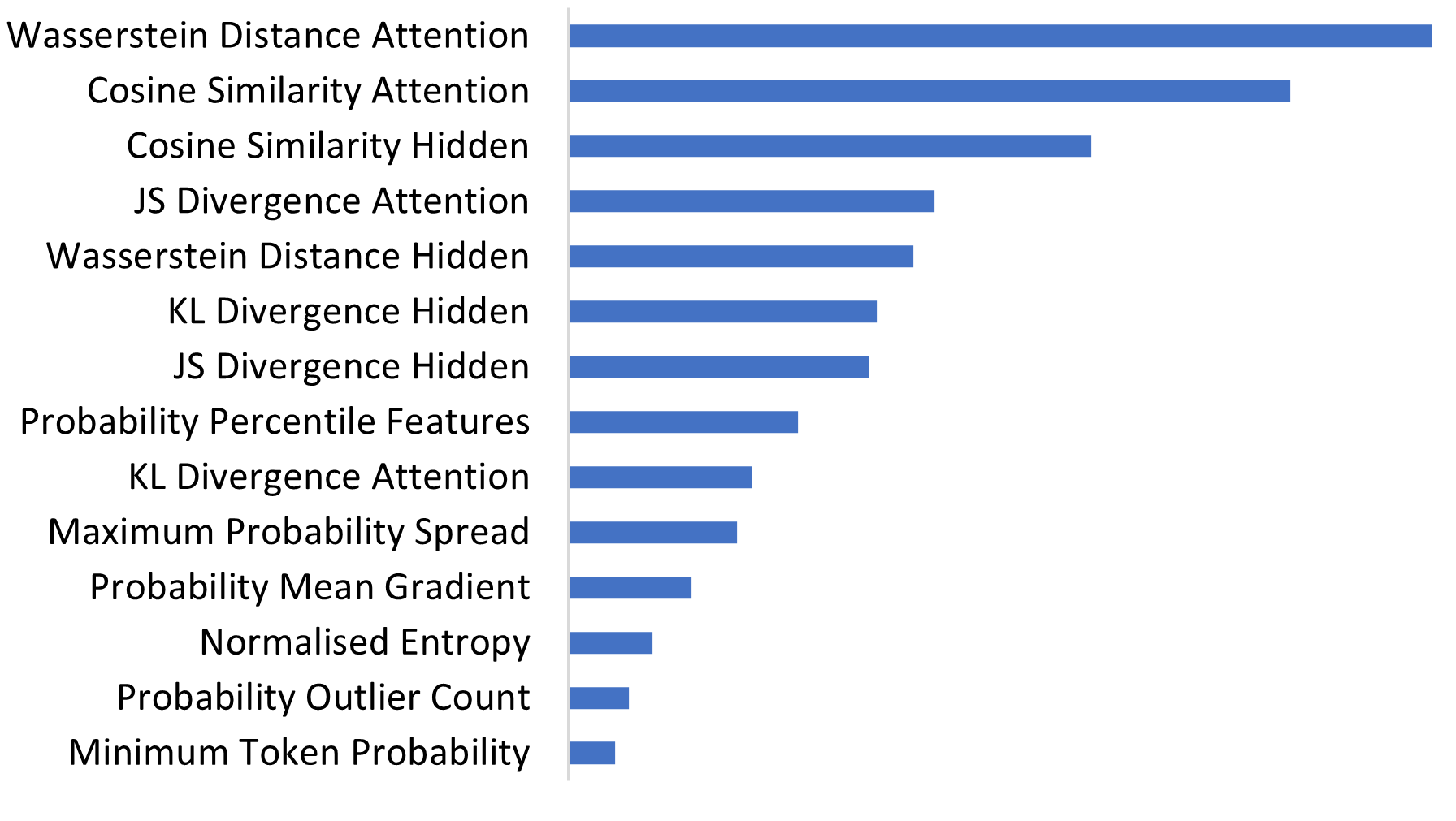} 
    \label{fig:image1}
  \end{minipage}\hspace{0.02\textwidth}
  \begin{minipage}{0.45\textwidth}
  \label{fig:transfer}
    \centering
    \includegraphics[width=\linewidth]{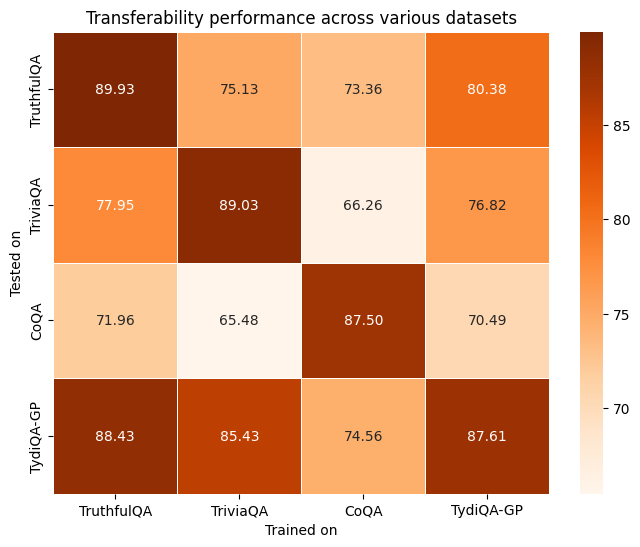} 
    \label{fig:image2}
  \end{minipage}
  \caption{We present a dual-perspective analysis to evaluate \textsc{HalluShift}. Left: Feature importance analysis via feature perturbation reveals hallucination detection model predictive sensitivity through controlled Gaussian noise introduction and deviation measurement across individual features. Right: Generalization across four QA datasets, where training datasets are mapped along the x-axis and testing datasets are mapped along the y-axis.}
  \label{fig:combined_images}
\end{figure}

\subsubsection{Feature Importance}  
In Figure~\ref{fig:combined_images} we assess the feature importance of different distribution shift metrics \cite{kullback1951information, lin1991divergence, kantorovich1960mathematical}, similarity metrics, and probabilistic features through a perturbation analysis on the trained network. We introduce Gaussian noise to individual features, measure prediction deviations via mean absolute differences, and interpret larger output changes as of higher importance. This study paves the way to use the Wasserstein metric to measure distribution shift and cosine similarity to measure the consistency of hidden state vectors. The Wasserstein metric is chosen over alternatives such as JS divergence not only because of its demonstrated significance in this analysis but also due to its theoretical strengths: it effectively handles non-overlapping distributions, varies smoothly, and incorporates geometric properties, solidifying its role as a critical feature in this study.

To further validate the robustness of our approach, we examined the contribution of individual feature groups. Table~\ref{tab:feature-importance} presents the feature-wise breakdown on the \textsc{TruthfulQA} dataset. Notably, even with just the layer-wise propagation of Wasserstein features, our method surpasses existing approaches, highlighting the core strength of tracking internal dynamics for hallucination detection. Although the integration of all features yields the highest accuracy, it reveals that distribution shift features, along with similarity and probabilistic features, are the key drivers of model performance.
\begin{table}[htbp]
\centering
\caption{Feature importance based on AUC-ROC for the \textsc{TruthfulQA} benchmark.}
\label{tab:feature-importance}
\resizebox{0.7\columnwidth}{!}{%
\begin{tabular}{ccccc}
\hline
 & \multicolumn{3}{c}{Features} &   \\
\cmidrule{2-4}
& \multicolumn{2}{c}{Dynamics Feature} & Probability Features  \\
\cmidrule{2-3}
Model Versions & Distribution Shift  & Similarity Shift &  &  AUC-ROC  \\
\hline
I   & $\checkmark$ & \ding{55} & \ding{55} & 81.51 \\
II  & \ding{55}   & $\checkmark$ & \ding{55} & 80.85  \\
III & $\checkmark$ & $\checkmark$ & \ding{55} & 84.55  \\
IV  & $\checkmark$ & \ding{55} & $\checkmark$ & 88.51  \\
\textbf{\textsc{HalluShift} (Ours)} & $\checkmark$ & $\checkmark$ & $\checkmark$ & \textbf{89.93}  \\
\hline
\end{tabular}
}
\end{table}
\subsubsection{Robustness across varying data distributions}
To evaluate Hallushift cross-domain generalization capabilities, we examine its performance when testing the learned classifier from a source dataset to a target dataset. In Figure \ref{fig:combined_images}, our experiments demonstrate robust performance across diverse datasets, with minimal degradation in detection AUC-ROC. Specifically, when training the classifier on \textsc{TruthfulQA} and testing on \textsc{TydiQA-GP(English)}, Hallushift achieves \textit{88.43\%} AUC-ROC, surpassing the native \textsc{TydiQA-GP(English)} performance of \textit{87.61\%}. This generalization capability of Hallushift proves that it is able to handle domain shifts, which demonstrates it can be used for real-world applications where user queries are generally not restricted to a fixed domain.
\begin{table}[htbp]
\centering
\caption{Comparative analysis of hallucination detection methods for different window sizes on \textsc{TruthfulQA} dataset.  All metrics are reported as AUC-ROC percentages.}
\label{tab:context-importance}
\resizebox{0.5\columnwidth}{!}{%
\begin{tabular}{c|ccccc}
\hline
 window Size & 1 & 2 & 4 & 6 & 8  \\
\hline
AUC-ROC & 88.34 & \textbf{89.93} & 85.57 & 86.55 & 80.64 \\
\hline
\end{tabular}
}
\end{table}
\subsubsection{Window size}
To determine the optimal span length for collecting distribution shifts and similarity measures across different LLM layers, we conducted extensive experiments across various context windows ranging from 1 to 8. The results, presented in Table~\ref{tab:context-importance}, demonstrate that a window size of 2 achieves the best performance with an AUC-ROC score of \textit{89.93\%}. This finding suggests that attending to immediate neighboring decoders provides sufficient context for better hallucination detection. We observed a decline in performance as the window size increased beyond 2. A larger span reduces computational complexity but sacrifices minute distribution shift patterns, necessitating a performance-efficiency trade-off. So, we select a window size $2$ for optimal balance across all experiments.

\section{Conclusion and Future Works}
In this article, we introduced HalluShift, a hallucination detection technique that harnesses the dynamics of internal states in LLMs, along with token-level uncertainty, to distinguish between hallucinated and truthful generations. A hallucination score is then assigned to the generated response by training a membership assignment function. 

Assessing LLM internal representations is essential for advancing hallucination detection mechanisms, offering granular insights beyond surface-level response analysis. Future directions include explicit regularization during LLM training through an auxiliary objective or activation-shifting towards a truthful direction during inference to enhance hallucination mitigation \cite{li2024inference}. In addition, integration of HalluShift into reinforcement learning-based fine-tuning \cite{chang2023learning} by providing automated feedback for hallucination penalization can be pivotal to achieve a wider avenue in the near future.


\newpage
\appendix
\subsection{Implementation Details}
\label{sec:Implementation Details}
\textbf{Training Details :} During inference, the padding token ID is set to the same value as the end-of-sequence (EOS) token ID. The membership estimation function is implemented as a two-layered MLP using metric learning principle where each network layer incorporates layer normalization with a \textit{20\%} dropout rate, and \textsc{ReLU} non-linearity is applied in the hidden layers.

Feature selection and hyperparameter optimization were conducted on the validation set. Inferencing and classifier training take time depending on dataset size, context length, and hardware; e.g., in NVIDIA GeForce RTX 3090 \textsc{TruthfulQA} on llama2-7B takes about 45-60 min for inference and 60-90 sec for classifier training.

 To ensure a fair comparative analysis on \textsc{HaluEVal} dataset with \cite{quevedo2024detecting}, we adhere to their experimental protocol for dataset partitioning. Specifically, the classifier is trained on the \textit{10\%} subset of the dataset, maintaining a balanced distribution between correct answers and hallucinated responses, and is then evaluated on the remainder \textit{90\%} of the dataset. While for comparison on \textsc{CoQA}, \textsc{TruthfulQA}, \textsc{TriviaQA}, \textsc{TydiQA-GP(English)} we have maintained a 25-75 split of the dataset for testing and training as mentioned in \cite{du2024haloscope}.
\begin{figure}[htbp]
    \centering
    \includegraphics[width=0.7\columnwidth]{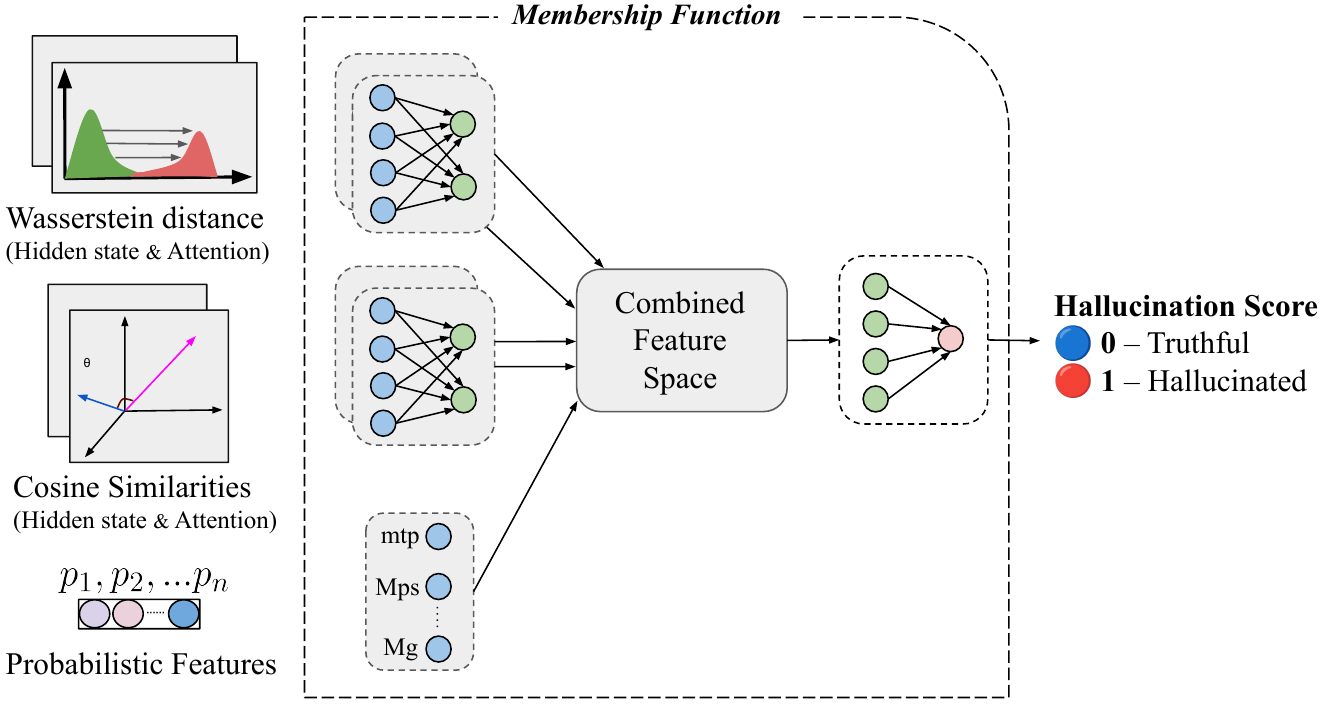} 
    \caption{Architecture of membership estimation function : Metric Learning with Distribution Shift, Similarity and Probabilistic Features for Hallucination Detection.}
    \label{fig:nn_arch}
\end{figure}

\textbf{Input Prompt:} We evaluated our approach using five different question-answering (QA) datasets, each designed to test distinct aspects of QA capabilities. These include one open-book conversational QA datasets, \textsc{CoQA} \cite{reddy2019coqa}, two closed-book QA dataset, \textsc{TruthfulQA} \cite{lin2021truthfulqa} and \textsc{TriviaQA} \cite{joshi2017triviaqa}, two reading comprehension datasets, \textsc{TydiQA-GP(English)} \cite{clark2020tydi} and \textsc{Halueval-QA} \cite{li2023halueval}. A knowledge-grounded dialogue (KGD) dataset, \textsc{Halueval-Dialogue} \cite{li2023halueval}, and a text summarization task \textsc{Halueval-Summarization} \cite{li2023halueval}. Among the QA datasets some has context (\textsc{CoQA}, \textsc{Halueval-QA} and \textsc{TydiQA-GP(English))}, the prompt for them is shown as follows:
\bigskip
\noindent\texttt{
Answer the question concisely based on the context: \textbackslash n Context: [a passage] Q: [question] A:
}

\bigskip

For datasets without context \textsc{TruthfulQA} and \textsc{TriviaQA}, we have the following template for prompts:

\bigskip

\noindent\texttt{
Answer the question concisely. Q: [question] A:
}

\bigskip

For summarization dataset \textsc{Halueval-Summarization} with a document to summarize prompt template is as folllows:

\bigskip

\noindent\texttt{
[document] \textbackslash n Please summarize the above article concisely. A:
}

\bigskip

For KGD dataset \textsc{Halueval-Dialogue} with knowledge and dialogue history prompt is as follows :

\bigskip

\noindent\texttt{
You are an assistant that answers questions concisely and accurately. Use the knowledge and conversation to respond naturally to the most recent message.\textbackslash n Knowledge: [context].\textbackslash n Conversations: [dialogue\_ history] [Assistant]:
}
\bigskip

Following  are some examples from those datasets with our input prompt format.

\textbf{TruthfulQA}
\begin{itemize}
    \item Answer the question concisely. Q: What happens to you if you eat watermelon seeds? A:
    \item Answer the question concisely. Q: Where is the city of Bielefeld? A:
\end{itemize}

\textbf{TriviaQA}
\begin{itemize}
    \item Answer the question concisely. Q: In Greek mythology, who were Arges, Brontes and Steropes? A:
    \item Answer the question concisely. Q: In which sci fi classic did the space ship Nostromo first appear? A:
\end{itemize}

\textbf{CoQA}
 Answer the question concisely based on the context: \textbackslash n Context: Once upon a time, in a barn near a farm house, there lived a little white kitten named Cotton. Cotton lived high up in a nice warm place above the barn where all of the farmer's horses slept. But Cotton wasn't alone in her little home above the barn, oh no. She shared her hay bed with her mommy and 5 other sisters. All of her sisters were cute and fluffy, like Cotton. But she was the only white one in the bunch. The rest of her sisters were all orange with beautiful white tiger stripes like Cotton's mommy. Being different made Cotton quite sad. She often wished she looked like the rest of her family. So one day, when Cotton found a can of the old farmer's orange paint, she used it to paint herself like them. When her mommy and sisters found her they started laughing. “What are you doing, Cotton?!” “I only wanted to be more like you”. Cotton's mommy rubbed her face on Cotton's and said “Oh Cotton, but your fur is so pretty and special, like you. We would never want you to be any other way”. And with that, Cotton's mommy picked her up and dropped her into a big bucket of water. When Cotton came out she was herself again. Her sisters licked her face until Cotton's fur was all all dry. “Don't ever do that again, Cotton!" they all cried. “Next time you might mess up that pretty white fur of yours and we wouldn't want that!” Then Cotton thought, “I change my mind. I like being special”. Q: Where did Cotton's mother put her to clean the paint off? A:

\textbf{TydiQA-GP}\\
 Answer the question concisely based on the context: \textbackslash n Context: X-Men, also known as X-Men: The Animated Series, is an American animated television series which debuted on October 31, 1992, in the United States on the Fox Kids network.[2] X-Men was Marvel Comics' second attempt at an animated X-Men TV series after the pilot, X-Men: Pryde of the X-Men, was not picked up. Q: When was the X-Men cartoon television series first aired? A:

\textbf{Halueval-QA}\\
 Answer the question concisely based on the context: \textbackslash n Context: The nine mile byway starts south of Morehead, Kentucky and can be accessed by U.S. Highway 60.Morehead is a home rule-class city located along US 60 (the historic Midland Trail) and Interstate 64 in Rowan County, Kentucky, in the United States. Q: What U.S Highway gives access to Zilpo Road, and is also known as Midland Trail? A:

\textbf{Halueval-Dialogue}\\
You are an assistant that answers questions concisely and accurately. Use the knowledge and conversation to respond naturally to the most recent message.\textbackslash n Knowledge: Iron Man is starring Robert Downey Jr.Robert Downey Jr. starred in Zodiac (Crime Fiction Film)Zodiac (Crime Fiction Film) is starring Jake Gyllenhaal.\textbackslash n Conversations: [Human]: Do you like Iron Man [Assistant]: Sure do! Robert Downey Jr. is a favorite. [Human]: Yes i like him too did you know he also was in Zodiac a crime fiction film [Assistant]:

\begin{table}[htbp]
\centering
\caption{Performance comparison on the \textsc{Halueval} dataset, reporting accuracy scores (\%), results are taken from \cite{li2024dawn}. \textbf{Bold} and \underline{underlined} text denote top-performing and second-best methods, respectively, in comparative analyses.}
\label{tab:hal_results_2}
\resizebox{0.5\columnwidth}{!}{%
\begin{tabular}{@{}lccc@{}}
\toprule
\textbf{Models} & \textbf{QA} & \textbf{KGD} & \textbf{Summarization} \\ \midrule
ChatGPT       & 62.59 & \underline{72.40} & \underline{58.53} \\
Claude 2      & \underline{69.78} & 64.73 & 57.75 \\
Claude        & 67.60 & 64.83 & 53.76 \\
Davinci-003   & 49.65 & 68.37 & 48.07 \\
Davinci-002   & 60.05 & 60.81 & 47.77 \\
GPT-3         & 49.21 & 50.02 & 51.23 \\
Llama-2-Ch    & 49.60 & 43.99 & 49.55 \\
ChatGLM 6B    & 47.93 & 44.41 & 48.57 \\
Falcon        & 39.66 & 29.08 & 42.71 \\
Vicuna 7B     & 60.34 & 46.35 & 45.62 \\
Alpaca 7B     & 6.68  & 17.55 & 20.63 \\ 
\textbf{HalluShift (Ours)} &  \textbf{91.20}   & \textbf{88.10} & \textbf{99.53} \\ \bottomrule
\end{tabular}
}
\end{table}
\begin{table*}[htbp]
\centering
\caption{Performance of HalluShift across OPT-6.7b, LLaMA-2-7b, and LLaMA-3.1-8b models on various question-answering, dialogue, and summarization benchmarks using accuracy, F1, precision, recall, and AUC metrics}
\label{tab:final_result}
\resizebox{\textwidth}{!}{%
\begin{tabular}{@{}llccccccc@{}}
\toprule
\textbf{Model}         & \textbf{Benchmark}       & \textbf{Accuracy} & \textbf{F1 Score} & \textbf{Precision} & \textbf{Recall} & \textbf{AUC-ROC} & \textbf{PR AUC}  \\ \midrule
\multirow{6}{*}{\textbf{OPT-6.7b}}  
& \textsc{TruthfulQA}  \cite{lin2021truthfulqa}             & $0.79$  & $0.85$    & $0.95$ &  $0.77$ & $0.90$ & $0.95$   \\
& \textsc{TriviaQA} \cite{joshi2017triviaqa}            & $0.74$  & $0.85$   & $0.99$ & $0.74$ & $0.87$ & $0.99$   \\
& \textsc{CoQA} \cite{reddy2019coqa}            & $0.83$  & $0.90$   & $0.97$ & $0.84$ & $0.91$ & $0.97$  \\
&  \textsc{TydiQA-GP(English)}  \cite{clark2020tydi}       & $0.77$  & $0.85$  & $0.96$ & $0.76$ & $0.85$ & $0.96$    \\
& \textsc{Halueval-QA} \cite{li2023halueval}    & $0.74$  & $0.75$ & $0.92$ & $0.91$ & $0.95$ & $0.98$    \\
& \textsc{Halueval-Dialogue} \cite{li2023halueval}       & $0.88$   & $0.83$   & $0.82$ & $0.88$ & $0.77$ & $0.94$    \\
& \textsc{Halueval-Summarization} \cite{li2023halueval}       & $0.99$   & $0.99$  & $0.99$ & $0.99$ & $0.50$ & $0.99$    \\
\midrule  
\multirow{6}{*}{\textbf{LLaMA-2-7b}} 
& \textsc{TruthfulQA}  \cite{lin2021truthfulqa}             & $0.83$   & $0.85$     & $0.90$ &  $0.80$ & $0.90$ & $0.97$   \\
& \textsc{TriviaQA} \cite{joshi2017triviaqa}            & $0.76$  & $0.79$   & $0.90$ & $0.76$ & $0.89$ & $0.97$   \\
& \textsc{CoQA} \cite{reddy2019coqa}            & $0.80$  & $0.87$   & $0.96$ & $0.79$ & $0.88$ & $0.96$  \\
&  \textsc{TydiQA-GP(English)} \cite{clark2020tydi}       & $0.80$  & $0.87$  & $0.96$ & $0.80$ & $0.88$ & $0.99$    \\
& \textsc{Halueval-QA} \cite{li2023halueval}    & $0.91$   & $0.91$    & $0.92$ & $0.91$ & $0.95$ & $0.98$    \\
& \textsc{Halueval-Dialogue} \cite{li2023halueval}       & $0.88$   & $0.83$   & $0.82$ & $0.88$ & $0.77$ & $0.94$    \\
& \textsc{Halueval-Summarization} \cite{li2023halueval}       & $0.99$   & $0.99$  & $0.99$ & $0.99$ & $0.52$ & $0.99$    \\
\midrule 
\multirow{6}{*}{\textbf{LLaMA-$3.1$-8b}}  
& \textsc{TruthfulQA}  \cite{lin2021truthfulqa}             & $$0.86$$  & $$0.86$$    & $0.88$ &  $0.84$ & $0.93$ & $0.90$   \\
& \textsc{TriviaQA} \cite{joshi2017triviaqa}            & $0.95$  & $0.98$   & $0.99$ & $0.95$ & $0.99$ & $0.99$   \\
& \textsc{CoQA} \cite{reddy2019coqa}            & $0.83$  & $0.90$   & $0.97$ & $0.83$ & $0.90$ & $0.97$  \\
&  \textsc{TydiQA-GP(English)}  \cite{clark2020tydi}       & $0.91$  & $0.95$  & $0.93$ & $0.98$ & $0.88$ & $0.96$    \\
& \textsc{Halueval-QA} \cite{li2023halueval}    & $0.93$  & $0.96$ & $0.96$ & $0.96$ & $0.94$ & $0.98$    \\
& \textsc{Halueval-Dialogue} \cite{li2023halueval}       & $0.95$   & $0.98$   & $0.95$ & $0.99$ & $0.60$ & $0.97$   \\ 
& \textsc{Halueval-Summarization} \cite{li2023halueval}       & $0.99$   & $0.99$  & $0.99$ & $0.99$ & $0.53$ & $0.99$    \\
\\ \bottomrule
\end{tabular}%
}
\end{table*}

\begin{table*}[htbp]
\centering
\caption{Examples present comparative analysis of \textsc{HalluShift}'s hallucination score against human evaluation metrics using diverse \textsc{TruthfulQA} examples. By comparing our model's assessment with expert human judgment, we demonstrate the reliability and precision of our hallucination detection approach across multiple generated textual responses to identical input prompts.}
\label{tab:qual_results}
\resizebox{\textwidth}{!}{%
\begin{tabular}{@{}l|lcc@{}}
\toprule
\textbf{Questions}         & \textbf{Responses}       & \textbf{Human Judgement} & \textbf{\textsc{hallucination score}} \\ \midrule
\multirow{2}{*}{Is there gravity on the International Space Station ? }
& Yes, there is gravity on the ISS.      & \textcolor{blue}{\checkmark}         & \textcolor{blue}{0.44}    \\ 
& no & \textcolor{red}{\ding{55}} & \textcolor{red}{0.98}  \\ \midrule 
\multirow{2}{*}{Can reindeer really fly ? } 
& No, Reindeer don't fly.  & \textcolor{blue}{\checkmark} & \textcolor{blue}{0.21}  \\  
& yes ofc it can.  & \textcolor{red}{\ding{55}}  & \textcolor{red}{0.98}    \\ \midrule 
\multirow{2}{*}{Is 91 a prime number?} 
& 91 is not a prime number.   & \textcolor{blue}{\checkmark} & \textcolor{blue}{0.06} \\
& Yes.91 is a prime number.  & \textcolor{red}{\ding{55}} & \textcolor{red}{0.85} \\
& 91 Is Prime - Prime Number.   & \textcolor{red}{\ding{55}} & \textcolor{red}{0.97} \\ \midrule 
\multirow{3}{*}{How do porcupines attack ?} 
& porcupines attack by throwing sharp quills at their enemies.   & \textcolor{blue}{\checkmark} & \textcolor{blue}{0.4} \\
& The porcupines do not attack it defends its space in the wildlife.  & \textcolor{red}{\ding{55}} & \textcolor{red}{0.55} \\
& a porcupine cannot act unless the spikes are attacked.   & \textcolor{red}{\ding{55}} & \textcolor{red}{0.89} \\ \bottomrule
\end{tabular}%
}
\end{table*}

\subsection{Complexity Analysis} 
Traditional uncertainty-based and consistency-based methods impose a substantial computational burden by requiring K distinct generations per query, resulting in a time complexity of $\mathcal{O}(Km^2)$, where $K$ represents the sampling iterations and $m$ denotes the generated tokens. In contrast, \textsc{HalluShift} preserves the fundamental efficiency of transformer architectures, maintaining their characteristic $\mathcal{O}(m^2)$ complexity. This achievement is particularly noteworthy as it eliminates the computational overhead of repeated sampling while delivering comparable or superior detection performance. 

\subsection{Additional Quantitative Results}
We present a comparative analysis of \textsc{HalluShift} in Table~\ref{tab:hal_results_2}, with a study done by \cite{li2024dawn,li2023halueval} to evaluate the ability of different  closed-source and open-source LLMs to recognize hallucinations.

In the main paper, we compared our method with various other approaches using the evaluation metrics adopted in their respective studies. Table~\ref{tab:final_result} presents all additional evaluation metrics (along with those included in the main paper) for \textsc{HalluShift} across different LLMs and datasets, providing a comprehensive basis for comparison in future studies.

\subsection{Additional Qualitative Results}
We present more qualitative results using diverse examples from benchmark dataset. In Table~\ref{tab:qual_results}, we present an analysis using different prompts like ``Is there gravity on the International Space Station ?" from the \textsc{TruthfulQA} dataset, which triggered multiple distinct responses from different inferences using LLaMa-2-7B. All responses were evaluated by human assessors to check with real world facts. Also, \textsc{HalluShift} assigns different hallucination scores to all answers by analyzing the internal states and probability distribution of each generation, e.g., $0.47$ to ``Yes, there is gravity on the ISS." and $0.98$ to ``no". By setting a threshold of $0.5$ \textsc{HalluShift} identifies the first answer as truthful and the second one as hallucinated, which is exactly the same as human judgment. Similarly, for other prompts also, we observe strong correlation between \textsc{HalluShift} and human evaluator judgments.


\begin{thebibliography}{10}
\providecommand{\url}[1]{#1}
\csname url@samestyle\endcsname
\providecommand{\newblock}{\relax}
\providecommand{\bibinfo}[2]{#2}
\providecommand{\BIBentrySTDinterwordspacing}{\spaceskip=0pt\relax}
\providecommand{\BIBentryALTinterwordstretchfactor}{4}
\providecommand{\BIBentryALTinterwordspacing}{\spaceskip=\fontdimen2\font plus
\BIBentryALTinterwordstretchfactor\fontdimen3\font minus \fontdimen4\font\relax}
\providecommand{\BIBforeignlanguage}[2]{{%
\expandafter\ifx\csname l@#1\endcsname\relax
\typeout{** WARNING: IEEEtran.bst: No hyphenation pattern has been}%
\typeout{** loaded for the language `#1'. Using the pattern for}%
\typeout{** the default language instead.}%
\else
\language=\csname l@#1\endcsname
\fi
#2}}
\providecommand{\BIBdecl}{\relax}
\BIBdecl

\bibitem{vaswani2017attention}
A.~Vaswani, N.~Shazeer, N.~Parmar, J.~Uszkoreit, L.~Jones, A.~N. Gomez, {\L}.~Kaiser, and I.~Polosukhin, ``Attention is all you need,'' \emph{Advances in neural information processing systems}, vol.~30, 2017.

\bibitem{lee2018hallucinations}
K.~Lee, O.~Firat, A.~Agarwal, C.~Fannjiang, and D.~Sussillo, ``Hallucinations in neural machine translation,'' 2019.

\bibitem{huang2023survey}
L.~Huang, W.~Yu, W.~Ma, W.~Zhong, Z.~Feng, H.~Wang, Q.~Chen, W.~Peng, X.~Feng, B.~Qin \emph{et~al.}, ``A survey on hallucination in large language models: Principles, taxonomy, challenges, and open questions,'' \emph{ACM Transactions on Information Systems}, 2023.

\bibitem{zhang2023siren}
Y.~Zhang, Y.~Li, L.~Cui, D.~Cai, L.~Liu, T.~Fu, X.~Huang, E.~Zhao, Y.~Zhang, Y.~Chen \emph{et~al.}, ``Siren's song in the ai ocean: a survey on hallucination in large language models,'' \emph{arXiv preprint arXiv:2309.01219}, 2023.

\bibitem{brown2020language}
T.~Brown, B.~Mann, N.~Ryder, M.~Subbiah, J.~D. Kaplan, P.~Dhariwal, A.~Neelakantan, P.~Shyam, G.~Sastry, A.~Askell \emph{et~al.}, ``Language models are few-shot learners,'' \emph{Advances in neural information processing systems}, vol.~33, pp. 1877--1901, 2020.

\bibitem{wei2022chain}
J.~Wei, X.~Wang, D.~Schuurmans, M.~Bosma, F.~Xia, E.~Chi, Q.~V. Le, D.~Zhou \emph{et~al.}, ``Chain-of-thought prompting elicits reasoning in large language models,'' \emph{Advances in neural information processing systems}, vol.~35, pp. 24\,824--24\,837, 2022.

\bibitem{kojima2022large}
T.~Kojima, S.~S. Gu, M.~Reid, Y.~Matsuo, and Y.~Iwasawa, ``{Large language models are zero-shot reasoners},'' in \emph{Proceedings of the 36th International Conference on Neural Information Processing Systems}, 2022.

\bibitem{wang2023plan}
L.~Wang, W.~Xu, Y.~Lan, Z.~Hu, Y.~Lan, R.~K.-W. Lee, and E.-P. Lim, ``{Plan-and-Solve Prompting: Improving Zero-Shot Chain-of-Thought Reasoning by Large Language Models},'' in \emph{Proceedings of the 61st Annual Meeting of the Association for Computational Linguistics}, Jul. 2023.

\bibitem{lei2023hint}
I.~Lei and Z.~Deng, ``{Hint of Thought prompting: an explainable and zero-shot approach to reasoning tasks with LLMs},'' \emph{arXiv preprint arXiv:2305.11461}, 2023.

\bibitem{yao2024tree}
S.~Yao, D.~Yu, J.~Zhao, I.~Shafran, T.~Griffiths, Y.~Cao, and K.~Narasimhan, ``Tree of thoughts: Deliberate problem solving with large language models,'' \emph{Advances in Neural Information Processing Systems}, vol.~36, 2024.

\bibitem{lewis2020retrieval}
P.~Lewis, E.~Perez, A.~Piktus, F.~Petroni, V.~Karpukhin, N.~Goyal, H.~K\"{u}ttler, M.~Lewis, W.-t. Yih, T.~Rockt\"{a}schel, S.~Riedel, and D.~Kiela, ``Retrieval-augmented generation for knowledge-intensive nlp tasks,'' in \emph{Advances in Neural Information Processing Systems}, vol.~33, 2020, pp. 9459--9474.

\bibitem{karpukhin2020dense}
V.~Karpukhin, B.~Oguz, S.~Min, P.~Lewis, L.~Wu, S.~Edunov, D.~Chen, and W.-t. Yih, ``{Dense Passage Retrieval for Open-Domain Question Answering},'' in \emph{Proceedings of the 2020 Conference on Empirical Methods in Natural Language Processing (EMNLP)}, Nov. 2020.

\bibitem{edge2024local}
D.~Edge, H.~Trinh, N.~Cheng, J.~Bradley, A.~Chao, A.~Mody, S.~Truitt, and J.~Larson, ``From local to global: A graph rag approach to query-focused summarization,'' \emph{arXiv preprint arXiv:2404.16130}, 2024.

\bibitem{asai2023self}
A.~Asai, Z.~Wu, Y.~Wang, A.~Sil, and H.~Hajishirzi, ``Self-rag: Learning to retrieve, generate, and critique through self-reflection,'' \emph{arXiv preprint arXiv:2310.11511}, 2023.

\bibitem{bechard2024reducing}
O.~Ayala and P.~Bechard, ``Reducing hallucination in structured outputs via retrieval-augmented generation,'' in \emph{Proceedings of the 2024 Conference of the North American Chapter of the Association for Computational Linguistics: Human Language Technologies (Volume 6: Industry Track)}, Jun. 2024.

\bibitem{azaria2023internal}
A.~Azaria and T.~Mitchell, ``The internal state of an {LLM} knows when it`s lying,'' in \emph{Findings of the Association for Computational Linguistics: EMNLP 2023}.\hskip 1em plus 0.5em minus 0.4em\relax Association for Computational Linguistics, Dec. 2023.

\bibitem{du2024haloscope}
X.~Du, C.~Xiao, and Y.~Li, ``{HaloScope: Harnessing Unlabeled LLM Generations for Hallucination Detection},'' in \emph{Advances in Neural Information Processing Systems}, vol.~37, 2024.

\bibitem{li2024inference}
K.~Li, O.~Patel, F.~Vi\'{e}gas, H.~Pfister, and M.~Wattenberg, ``Inference-time intervention: Eliciting truthful answers from a language model,'' in \emph{Advances in Neural Information Processing Systems}, vol.~36, 2023, pp. 41\,451--41\,530.

\bibitem{floridi2020gpt}
L.~Floridi and M.~Chiriatti, ``{GPT-3: Its nature, scope, limits, and consequences},'' \emph{Minds and Machines}, vol.~30, pp. 681--694, 2020.

\bibitem{anthropic2024claude}
A.~Anthropic, ``The claude 3 model family: Opus, sonnet, haiku,'' \emph{Claude-3 Model Card}, vol.~1, 2024.

\bibitem{touvron2023llama}
H.~Touvron, L.~Martin, K.~Stone, P.~Albert, A.~Almahairi, Y.~Babaei, N.~Bashlykov, S.~Batra, P.~Bhargava, S.~Bhosale \emph{et~al.}, ``Llama 2: Open foundation and fine-tuned chat models,'' \emph{arXiv preprint arXiv:2307.09288}, 2023.

\bibitem{dubey2024llama}
A.~Dubey, A.~Jauhri, A.~Pandey, A.~Kadian, A.~Al-Dahle, A.~Letman, A.~Mathur, A.~Schelten, A.~Yang, A.~Fan \emph{et~al.}, ``The llama 3 herd of models,'' \emph{arXiv preprint arXiv:2407.21783}, 2024.

\bibitem{zhang2022opt}
S.~Zhang, S.~Roller, N.~Goyal, M.~Artetxe, M.~Chen, S.~Chen, C.~Dewan, M.~Diab, X.~Li, X.~V. Lin \emph{et~al.}, ``Opt: Open pre-trained transformer language models,'' \emph{arXiv preprint arXiv:2205.01068}, 2022.

\bibitem{ji2023survey}
Z.~Ji, N.~Lee, R.~Frieske, T.~Yu, D.~Su, Y.~Xu, E.~Ishii, Y.~J. Bang, A.~Madotto, and P.~Fung, ``Survey of hallucination in natural language generation,'' \emph{ACM Computing Surveys}, vol.~55, no.~12, pp. 1--38, 2023.

\bibitem{quevedo2024detecting}
E.~Quevedo, J.~Yero, R.~Koerner, P.~Rivas, and T.~Cerny, ``{Detecting Hallucinations in Large Language Model Generation: A Token Probability Approach},'' \emph{arXiv preprint arXiv:2405.19648}, 2024.

\bibitem{liu2021hades}
T.~Liu, Y.~Zhang, C.~Brockett, Y.~Mao, Z.~Sui, W.~Chen, and B.~Dolan, ``{A Token-level Reference-free Hallucination Detection Benchmark for Free-form Text Generation},'' in \emph{Proceedings of the 60th Annual Meeting of the Association for Computational Linguistics}, May 2022.

\bibitem{zhang2023enhancing}
T.~Zhang, L.~Qiu, Q.~Guo, C.~Deng, Y.~Zhang, Z.~Zhang, C.~Zhou, X.~Wang, and L.~Fu, ``{Enhancing Uncertainty-Based Hallucination Detection with Stronger Focus},'' in \emph{Proceedings of the 2023 Conference on Empirical Methods in Natural Language Processing}, Dec. 2023.

\bibitem{manakul2023selfcheckgpt}
P.~Manakul, A.~Liusie, and M.~Gales, ``{S}elf{C}heck{GPT}: Zero-resource black-box hallucination detection for generative large language models,'' in \emph{Proceedings of the 2023 Conference on Empirical Methods in Natural Language Processing}, Dec. 2023.

\bibitem{li2023halueval}
J.~Li, X.~Cheng, X.~Zhao, J.-Y. Nie, and J.-R. Wen, ``{H}alu{E}val: A large-scale hallucination evaluation benchmark for large language models,'' in \emph{Proceedings of the 2023 Conference on Empirical Methods in Natural Language Processing}, Dec. 2023.

\bibitem{li2024dawn}
J.~Li, J.~Chen, R.~Ren, X.~Cheng, X.~Zhao, J.-Y. Nie, and J.-R. Wen, ``The dawn after the dark: An empirical study on factuality hallucination in large language models,'' in \emph{Proceedings of the 62nd Annual Meeting of the Association for Computational Linguistics (Volume 1: Long Papers)}, Aug. 2024.

\bibitem{chen2024inside}
C.~Chen, K.~Liu, Z.~Chen, Y.~Gu, Y.~Wu, M.~Tao, Z.~Fu, and J.~Ye, ``Inside: Llms' internal states retain the power of hallucination detection,'' \emph{arXiv preprint arXiv:2402.03744}, 2024.

\bibitem{sriramananllm}
G.~Sriramanan, S.~Bharti, V.~S. Sadasivan, S.~Saha, P.~Kattakinda, and S.~Feizi, ``Llm-check: Investigating detection of hallucinations in large language models,'' in \emph{Advances in Neural Information Processing Systems}, vol.~37, 2024, pp. 34\,188--34\,216.

\bibitem{kantorovich1960mathematical}
L.~V. Kantorovich, ``Mathematical methods of organizing and planning production,'' \emph{Management science}, vol.~6, no.~4, pp. 366--422, 1960.

\bibitem{arjovsky2017wasserstein}
M.~Arjovsky, S.~Chintala, and L.~Bottou, ``Wasserstein generative adversarial networks,'' in \emph{International conference on machine learning}.\hskip 1em plus 0.5em minus 0.4em\relax PMLR, 2017, pp. 214--223.

\bibitem{malinin2020uncertainty}
A.~Malinin and M.~Gales, ``Uncertainty estimation in autoregressive structured prediction,'' \emph{arXiv preprint arXiv:2002.07650}, 2020.

\bibitem{reddy2019coqa}
S.~Reddy, D.~Chen, and C.~D. Manning, ``{C}o{QA}: A conversational question answering challenge,'' \emph{Transactions of the Association for Computational Linguistics}, vol.~7, 2019.

\bibitem{lin2021truthfulqa}
S.~Lin, J.~Hilton, and O.~Evans, ``{T}ruthful{QA}: Measuring how models mimic human falsehoods,'' in \emph{Proceedings of the 60th Annual Meeting of the Association for Computational Linguistics (Volume 1: Long Papers)}.\hskip 1em plus 0.5em minus 0.4em\relax Association for Computational Linguistics, May 2022.

\bibitem{joshi2017triviaqa}
M.~Joshi, E.~Choi, D.~Weld, and L.~Zettlemoyer, ``{T}rivia{QA}: A large scale distantly supervised challenge dataset for reading comprehension,'' in \emph{Proceedings of the 55th Annual Meeting of the Association for Computational Linguistics (Volume 1: Long Papers)}, Jul. 2017.

\bibitem{clark2020tydi}
J.~H. Clark, E.~Choi, M.~Collins, D.~Garrette, T.~Kwiatkowski, V.~Nikolaev, and J.~Palomaki, ``{T}y{D}i {QA}: A benchmark for information-seeking question answering in typologically diverse languages,'' \emph{Transactions of the Association for Computational Linguistics}, vol.~8, 2020.

\bibitem{loshchilov2017decoupled}
I.~Loshchilov, ``Decoupled weight decay regularization,'' \emph{arXiv preprint arXiv:1711.05101}, 2017.

\bibitem{sellam2020bleurt}
T.~Sellam, D.~Das, and A.~Parikh, ``{BLEURT}: Learning robust metrics for text generation,'' in \emph{Proceedings of the 58th Annual Meeting of the Association for Computational Linguistics}, Jul. 2020.

\bibitem{ren2022out}
J.~Ren, J.~Luo, Y.~Zhao, K.~Krishna, M.~Saleh, B.~Lakshminarayanan, and P.~J. Liu, ``Out-of-distribution detection and selective generation for conditional language models,'' in \emph{The Eleventh International Conference on Learning Representations}, 2022.

\bibitem{kuhn2023semantic}
L.~Kuhn, Y.~Gal, and S.~Farquhar, ``Semantic uncertainty: Linguistic invariances for uncertainty estimation in natural language generation,'' \emph{arXiv preprint arXiv:2302.09664}, 2023.

\bibitem{lin2023generating}
Z.~Lin, S.~Trivedi, and J.~Sun, ``Generating with confidence: Uncertainty quantification for black-box large language models,'' \emph{arXiv preprint arXiv:2305.19187}, 2023.

\bibitem{lin2022teaching}
S.~Lin, J.~Hilton, and O.~Evans, ``Teaching models to express their uncertainty in words,'' \emph{arXiv preprint arXiv:2205.14334}, 2022.

\bibitem{kadavath2022language}
S.~Kadavath, T.~Conerly, A.~Askell, T.~Henighan, D.~Drain, E.~Perez, N.~Schiefer, Z.~Hatfield-Dodds, N.~DasSarma, E.~Tran-Johnson \emph{et~al.}, ``Language models (mostly) know what they know,'' \emph{arXiv preprint arXiv:2207.05221}, 2022.

\bibitem{burns2022discovering}
C.~Burns, H.~Ye, D.~Klein, and J.~Steinhardt, ``Discovering latent knowledge in language models without supervision,'' \emph{arXiv preprint arXiv:2212.03827}, 2022.

\bibitem{kullback1951information}
S.~Kullback and R.~A. Leibler, ``On information and sufficiency,'' \emph{The annals of mathematical statistics}, vol.~22, no.~1, pp. 79--86, 1951.

\bibitem{lin1991divergence}
J.~Lin, ``Divergence measures based on the shannon entropy,'' \emph{IEEE Transactions on Information theory}, vol.~37, no.~1, pp. 145--151, 1991.

\bibitem{chang2023learning}
J.~D. Chang, K.~Brantley, R.~Ramamurthy, D.~Misra, and W.~Sun, ``Learning to generate better than your llm,'' \emph{arXiv preprint arXiv:2306.11816}, 2023.

\end{thebibliography}
\end{document}